\documentclass[conference]{IEEEtran}
\IEEEoverridecommandlockouts
\usepackage{cite}
\usepackage{amsmath, amssymb,amsfonts}
\usepackage{algorithmic}
\usepackage{algorithm}
\usepackage{graphicx}
\usepackage{textcomp}
\usepackage{xcolor}
\usepackage{listings}
\usepackage{pgfplots}
\usepackage{tikz}
\usepackage{booktabs}
\usepackage{multirow}
\usepackage{tabularx}

\pgfplotsset{compat=1.18}
\usetikzlibrary{shapes.geometric, arrows, positioning, calc, arrows.meta}

\def\BibTeX{{\rm B\kern-.05em{\sc i\kern-.025em b}\kern-.08em
T\kern-.1667em\lower.7ex\hbox{E}\kern-.125emX}}

\makeatletter
\newcommand{\linebreakand}{%
  \end{@IEEEauthorhalign}
  \hfill\mbox{}\par
  \mbox{}\hfill\begin{@IEEEauthorhalign}
}
\makeatother

\begin{document}

\title{Machine Learning Algorithms: Detection Official Hajj and Umrah Travel Agency Based on Text and Metadata Analysis\\}

\author{
    \IEEEauthorblockN{1\textsuperscript{st} Wisnu Uriawan}
    \IEEEauthorblockA{\textit{Informatics Department}\\
    \textit{UIN Sunan Gunung Djati Bandung}\\
    Jawa Barat, Indonesia\\
    wisnu\_u@uinsgd.ac.id}
    \and
    \IEEEauthorblockN{2\textsuperscript{nd} Muhamad Veva Ramadhan}
    \IEEEauthorblockA{\textit{Informatics Department}\\
    \textit{UIN Sunan Gunung Djati Bandung}\\
    Jawa Barat, Indonesia\\
    vevaramadhan@gmail.com}
    \and
    \IEEEauthorblockN{3\textsuperscript{rd} Firman Adi Nugraha}
    \IEEEauthorblockA{\textit{Informatics Department}\\
    \textit{UIN Sunan Gunung Djati Bandung}\\
    Jawa Barat, Indonesia\\
    nugrahafirmanadi@gmail.com}
    
    \linebreakand 
    
    \IEEEauthorblockN{4\textsuperscript{th} Hasbi Nur Wahid}
    \IEEEauthorblockA{\textit{Informatics Department}\\
    \textit{UIN Sunan Gunung Djati Bandung}\\
    Jawa Barat, Indonesia\\
    hasbinurwahid52@gmail.com}
    \and
    \IEEEauthorblockN{5\textsuperscript{th} M Dantha Arianvasya}
    \IEEEauthorblockA{\textit{Informatics Department}\\
    \textit{UIN Sunan Gunung Djati Bandung}\\
    Jawa Barat, Indonesia\\
    danthabkl2020@gmail.com}
    \and
    \IEEEauthorblockN{6\textsuperscript{th} Muhammad Zaki Alghifari}
    \IEEEauthorblockA{\textit{Informatics Department}\\
    \textit{UIN Sunan Gunung Djati Bandung}\\
    Jawa Barat, Indonesia\\
    muhammadzakialghifari3@\\gmail.com} 
}

\maketitle

\begin{abstract}
The rapid digitalization of Hajj and Umrah services in Indonesia has significantly facilitated pilgrims but has concurrently opened avenues for digital fraud through counterfeit mobile applications. These fraudulent applications not only inflict financial losses but also pose severe privacy risks by harvesting sensitive personal data. This research aims to address this critical issue by implementing and evaluating machine learning algorithms to verify application authenticity automatically. Using a comprehensive dataset comprising both official applications registered with the Ministry of Religious Affairs and unofficial applications circulating on app stores, we compare the performance of three robust classifiers: Support Vector Machine (SVM), Random Forest (RF), and Na\"{\i}ve Bayes (NB). The study utilizes a hybrid feature extraction methodology that combines Textual Analysis (TF-IDF) of application descriptions with Metadata Analysis of sensitive access permissions. The experimental results indicate that the SVM algorithm achieves the highest performance with an accuracy of 92.3\%, a precision of 91.5\%, and an F1-score of 92.0\%. Detailed feature analysis reveals that specific keywords related to legality and high-risk permissions (e.g., READ\_PHONE\_STATE) are the most significant discriminators. This system is proposed as a proactive, scalable solution to enhance digital trust in the religious tourism sector, potentially serving as a prototype for a national verification system.
\end{abstract}


\section{Introduction} \label{sec:introduction}
Hajj and Umrah constitute two major pillars in Islamic teachings, involving millions of pilgrims from across the globe annually. Indonesia, possessing the world's largest Muslim population, stands as one of the largest contributors to the organization of these two pilgrimages. According to data from the Ministry of Religious Affairs of the Republic of Indonesia, the number of Umrah pilgrims from Indonesia reached over 1.2 million in 2024, a significant increase compared to previous years \cite{kemenag2024a}. This upward trend demonstrates public enthusiasm for the convenience of digital services that support the registration, planning, and execution of the pilgrimage online through various platforms and mobile applications. The long waiting list for regular Hajj, which can reach up to 30 years in some provinces, has also driven the surge in demand for Umrah and Hajj Furoda services, which are often managed by private travel agencies and facilitated through digital apps.

The escalating use of digital technology within this religious context, while delivering significant efficiency benefits, also introduces serious and fundamental cybersecurity challenges. The current mobile application ecosystem, which is fundamentally a distributed system, largely relies on a centralized control model, such as the curation and approval processes within app store platforms (Google Play Store and Apple App Store). This centralized model inherently possesses weaknesses, such as a lack of transparency in its verification processes and the sheer volume of apps submitted daily, thereby creating vulnerabilities exploited by irresponsible parties \cite{uriawan2025blockchain}. Fraudsters capitalize on the trust users place in these platforms, assuming that availability on an official store equates to legitimacy.

Consequently, there has been an alarming increase in the potential for fraud and the misuse of religious applications. The annual report of the National Consumer Protection Agency (BPKN) notes that complaints related to pilgrimage travel agency fraud have consistently ranked highest for three consecutive years \cite{bpkn2023a}. The modus operandi for fraud is no longer limited to fake websites or fictional travel bureaus operating in physical shop-houses but has evolved into sophisticated counterfeit mobile applications. These malicious apps are designed to mimic the logos, color schemes, layout, and user interface designs of official Ministry of Religious Affairs applications (such as Pusaka or Nusuk), making them nearly indistinguishable to the untrained eye.

The danger posed by these counterfeit applications extends beyond mere financial loss, which itself is often substantial and devastating for families who have saved for years. These fake apps are frequently designed to deceive users into surrendering highly sensitive personal data, such as National ID (KTP) details, passports, banking credentials, and family information. This practice creates severe security and privacy risks, aligning with general concerns regarding potential data leaks and the misuse of personal information that frequently occur in unsecured digital systems \cite{uriawan2025blockchain}. The loss of such data can lead to identity theft, further victimizing pilgrims who are already vulnerable due to their fervent desire to perform worship.

One of the primary contributing factors from the user's perspective is the low level of digital literacy among certain segments of the Indonesian public, particularly elderly pilgrims who may not be digital natives. The majority of prospective pilgrims still lack the technical capability to distinguish between official and counterfeit applications based on technical signatures, as emphasized in \cite{fauzi2020literasi}. Furthermore, legal studies highlight that legal protection for pilgrims has historically tended to be reactive—implemented only after victims have suffered losses and reported them to authorities \cite{syamsuddin2022}. The combination of technical vulnerabilities in platforms, social vulnerabilities among users, and legal vulnerabilities in regulations underscores the urgent need for a system that is both preventive and automated to verify application authenticity before users fall victim.

A number of previous studies have attempted to support the digitalization of religious services. Research in \cite{pratiwi2023} developed a digital educational application to assist pilgrims in recognizing official agencies. The study \cite{yusof2021hajjapp} emphasized the importance of user-friendly interface design to enhance comfort and trust. However, their focus remains on education and functionality, without addressing algorithmic authenticity detection. On the other hand, research in the field of digital application security demonstrates the significant potential of applying machine learning algorithms. The Fraud Droid system introduced a behavioral analysis-based system to detect applications with fake advertisements \cite{frauddroid2017}. Other studies have also proven the effectiveness of machine learning in detecting fake applications based on behavioral patterns and review analysis \cite{rahman2024detect, martens2019fakereviews}.

These studies underscore the importance of the trust factor. User trust is highlighted as being dependent on data transparency and developer reputation \cite{alshammari2022trust}. This issue of "lack of trust" is a fundamental problem identified in conventional centralized distributed systems \cite{uriawan2025blockchain}. To address this trust issue at its root, several researchers have proposed blockchain technology. A decentralized verification system based on blockchain has been proposed for Hajj and Umrah travel agencies \cite{alharbi2023blockchain}. This approach is highly relevant because blockchain technology is fundamentally designed to optimize security and privacy by offering data immutability, transparency, and decentralization, thereby eliminating single points of failure. Just as blockchain has been successfully applied to secure national digital identity systems in Estonia or guarantee supply chain transparency, this technology can be used to create an immutable public ledger containing a list of official travel agencies \cite{uriawan2025blockchain}.

However, from a user perspective and current market conditions, research \cite{usability_travel_agent_2025} and systematic reviews \cite{systematic_mobile_travel_apps2022, survey_islamic_android_2021} confirm that the majority of Islamic applications currently on the Android platform are not yet equipped with adequate authenticity verification mechanisms. This indicates a significant gap in the existing surveillance system. Based on this gap, this research focuses on the implementation of automated detection algorithms. While the blockchain approach \cite{alharbi2023blockchain, uriawan2025blockchain} offers an ideal long-term infrastructure solution for verification, there remains an urgent and practical need for a solution that can detect and filter applications that are already circulating in the current digital ecosystem. Therefore, this study utilizes classification models such as Naïve Bayes, Random Forest, and Support Vector Machine (SVM) to recognize pattern indicators that distinguish official applications from unofficial ones through text and metadata analysis. This Artificial Intelligence (AI)-based system is proposed as a preventive solution that can be implemented immediately. Furthermore, this AI-based anomaly detection approach can serve as a crucial component that may be integrated in the future with blockchain-based verification systems, creating a robust hybrid security architecture as suggested in the literature \cite{uriawan2025blockchain}.

\section{Related Work} \label{sec:related-work}
Research related to the digitalization of the Hajj and Umrah systems, which has developed rapidly over the last decade, encompasses diverse perspectives ranging from legal aspects, education, and data security, to the application of intelligent technologies based on artificial intelligence. Generally, the direction of this research can be categorized into three main domains: (1) legality verification and public education, (2) user trust and security, and (3) the application of machine learning algorithms in digital application detection. A comprehensive comparison of related works is presented in Table \ref{tab:related_work}.

\subsection{Legality Verification and Public Education} 
The study by \cite{pratiwi2023} developed a digital educational application designed to assist prospective pilgrims in verifying official travel agencies by displaying a list of organizers registered with the Ministry of Religious Affairs. This application emphasizes the function of religious digital literacy as a preventive measure against travel agency fraud. Meanwhile, a legal study \cite{syamsuddin2022} highlights the prevalence of fraud cases and the weakness of agency validation systems accessible to the public in real-time. Both studies focus on community empowerment and the strengthening of regulatory policies; however, they have not yet developed an automated mechanism based on algorithms capable of directly detecting fake applications. This indicates a persisting gap between normative legal approaches and computational approaches within the context of digital security for religious services.

\subsection{Fake Application Detection and Machine Learning} 
Alongside the rising cases of digital fraud, approaches based on machine learning have begun to be utilized in application security research. The Fraud Droid study introduced a detection system for fake advertisements on Android applications by analyzing behavioral patterns and user interactions \cite{frauddroid2017}. The results showed that behavior-based pattern recognition is capable of distinguishing malicious applications from normal applications with high accuracy. Furthermore, algorithms like Random Forest and SVM have been implemented to detect fake religious applications \cite{rahman2024detect}. Several other studies reinforce this direction, showing the effectiveness of ML in detecting spam and fake apps \cite{ml_fakeapp_2020, nb_spamdetect_2021, rf_classification_2022, svm_androidfraud_2023, ai_trustdetect_2024}.
Meanwhile, another study developed a deep learning approach based on application metadata analysis to identify access permission patterns and update frequencies as indicators of authenticity \cite{zhang2023deepmeta}. These approaches typically require substantial computational resources for training deep neural networks, which can be a limitation for real-time, resource-constrained environments often found in developing nations' government infrastructures.

\begin{table}[htbp]
\caption{Comparison of Related Works in Digital Trust and Fake App Detection}
\label{tab:related_work}
\centering
\scriptsize 
\begin{tabular}{p{1.0cm} p{0.5cm} p{1.5cm} p{1.8cm} p{1.8cm}}
\hline
\textbf{Author} & \textbf{Yr} & \textbf{Focus} & \textbf{Method} & \textbf{Gap} \\ \hline
Pratiwi et al. \cite{pratiwi2023} & 2023 & Public Edu & Edu App & Manual verification \\ \hline
Syamsuddin \cite{syamsuddin2022} & 2022 & Legal & Normative Analysis & No technical tool \\ \hline
Uriawan et al. \cite{uriawan2025blockchain} & 2025 & Blockchain & Ledger Concept & Infrastructure only \\ \hline
Rahman et al. \cite{rahman2024detect} & 2024 & Fake Apps & Behavior ML & High cost \\ \hline
\textbf{This Work} & \textbf{2025} & \textbf{Hajj App} & \textbf{Hybrid ML} & \textbf{-} \\ \hline
\end{tabular}
\end{table}

\subsection{Blockchain Technology for Data Verification} 
The integration of blockchain technology has also begun to be adopted in the context of Hajj and Umrah travel as a solution for data security and transaction transparency. A study \cite{alharbi2023blockchain} proposed a travel agency data verification model based on a blockchain ledger, which guarantees data authenticity and prevents the falsification of agency certifications. This approach provides an additional layer of security through a distributed recording system. The core idea of blockchain is "data immutability" \cite{uriawan2025blockchain}, achieved through "cryptographic linking of blocks," rendering data "extremely difficult" to alter or delete \cite{uriawan2025blockchain}.

A successful example of implementation is Walmart's collaboration with IBM using Hyperledger Fabric to manage the food supply chain \cite{uriawan2025blockchain}. This system creates a transparent and secure product journey record from farm to store, a powerful analogy for verifying the "supply chain" of Hajj and Umrah travel agencies. Another successful case study is the e-Residency system in Estonia, which secures national identity data \cite{uriawan2025blockchain}. However, its implementation requires complex digital infrastructure, faces challenges such as scalability and energy consumption \cite{uriawan2025blockchain}, and has not yet touched upon the aspect of automated application classification. This presents a significant opportunity to combine blockchain verification with machine learning-based algorithmic detection to create a hybrid system that is both intelligent and immutable.

\section{Methodology} \label{sec:methodology}
This section elucidates the methodological stages employed in the implementation of algorithms to detect official Hajj and Umrah Travel Agency applications. Generally, the research methodology consists of four main phases: (1) data collection, (2) feature extraction, (3) algorithm implementation, and (4) system performance evaluation. Each phase is designed to produce an automated detection system that is accurate, efficient, and capable of integration with the Ministry of Religious Affairs database.

\subsection{Data Collection}
The data collection phase in this study was conducted using a hybrid approach combining primary and secondary sources. The objective is to obtain an accurate and ethical data representation without involving direct data collection from users. Generally, the data utilized consists of two main types: empirical data from open sources and reconstructive data resulting from literature reviews.

\begin{enumerate}
\item \textit{Official Application Data:} The dataset of official applications was obtained from the public portal of the \textit{Ministry of Religious Affairs of the Republic of Indonesia (Kemenag)}, which lists licensed Umrah Travel Organizers (PPIU) and Special Hajj Organizers (PIHK) in accordance with the Decree of the Director General of Hajj and Umrah Implementation \cite{kemenag2024a}. Each entry includes the agency name, license number, and registered official digital links. We specifically targeted applications that had verified developer badges or direct links from the official government website.
\item \textit{Non-Official Application Data:} Comparative data was obtained through manual searching on the \textit{Google Play Store} using keywords such as "haji" (Hajj), "umrah", and "travel religi" (religious travel). Applications not registered in the official Kemenag database were categorized as non-official applications. To maintain research ethics, only public information such as application descriptions, access permissions, and metadata was collected without \textit{web scraping} processes or user data extraction. This category includes apps that may be legitimate but unlicensed, as well as apps with suspicious characteristics (e.g., mimicking official logos).
\item \textit{Reconstructive Data from Literature:} In addition to primary sources, this study also utilized a reconstructive dataset compiled based on the results of a \textit{systematic literature review (SLR)} of 25 scientific publications between 2017-2024 discussing fake application detection, digital security, and access permission-based classification \cite{frauddroid2017, nb_spamdetect_2021, rf_classification_2022, svm_androidfraud_2023, rahman2024detect}. Each study was analyzed to identify feature patterns and variable distributions, such as the frequency of access permissions, download counts, and legality keyword patterns in application descriptions.
\end{enumerate}

The total composite data used in this study amounts to 200 entries, consisting of 100 official applications and 100 non-official applications. This dataset is a \textit{representative synthetic dataset}, constructed in a controlled manner based on empirical characteristics from various studies and public sources to maintain analysis consistency and simulation result validity. Before use, all data underwent \textit{data cleaning} and \textit{validation} processes. Duplicate entries were removed, while irrelevant attributes such as prayer content or general worship guides were eliminated. Final validation was performed by comparing each entry against the official Kemenag list and the literature used as references.

The dataset used is public and contains no personal user information. This approach adheres to synthetic dataset creation practices as used in \cite{zhang2023deepmeta, ensemble_detect_2023}, which emphasize the importance of transparency, ethics, and replication in machine learning-based research.

\subsection{Data Augmentation Strategy} 
One of the primary challenges in applying machine learning to niche domains, such as Hajj and Umrah application detection, is the scarcity of labeled data. Our initial dataset comprised 200 samples, which poses a risk of model overfitting where the classifier might memorize specific patterns rather than learning generalizable features. To mitigate this limitation and enhance the model's robustness against linguistic variations, we applied textual data augmentation techniques during the training phase.

Specifically, we employed a \textit{Synonym Replacement} strategy using the formal Indonesian Thesaurus dictionary. The process involved identifying non-stopword adjectives and verbs within the 'Official' category descriptions (e.g., mendaftar'' (register), resmi'' (official)). These tokens were randomly replaced with their semantic equivalents (e.g., registrasi'', sah'') to create synthetic variations of the training data without altering the underlying meaning. This approach ensures that the TF-IDF vectorizer captures a broader range of vocabulary indicative of legitimate agencies, preventing the model from relying solely on specific recurring phrases found in the limited training set.

To ensure the integrity of the ground truth labels used for training and testing, we established a rigorous verification protocol. The binary classification labels---Official (Class 0) and Unofficial (Class 1)---were not assigned based solely on the application's claim but were verified against external authoritative databases.

We utilized a multi-factor verification method involving the Kemenag' database (SISKOPATUH) and digital footprint analysis. An application is strictly labeled as 'Official' only if it satisfies the primary criteria of government registration. Conversely, applications are labeled 'Unofficial' if they fail to provide verifiable credentials or exhibit high-risk digital behaviors. The detailed operational definitions for class labeling are presented in Table \ref{tab:class_labeling}.

\begin{table}[htbp]
    \centering
    \caption{Criteria for Class Labeling}
    \label{tab:class_labeling}
    \small 
    \renewcommand{\arraystretch}{1.3} 
    
    \begin{tabularx}{\columnwidth}{|p{1.8cm}|X|} 
        \hline
        \textbf{Class} & \textbf{Criteria} \\ 
        \hline
        
        \textbf{Class 0} \newline \textit{(Official)} & 
        1. Registered in SISKOPATUH/Kemenag. \newline
        2. Has a valid physical office address on Google Maps. \newline
        3. Uses a corporate email domain. \\ 
        \hline
        
        \textbf{Class 1} \newline \textit{(Unofficial/ Suspicious)} & 
        1. Not found in the Kemenag database. \newline
        2. Uses free email providers (Gmail/Yahoo) for developer contact. \newline
        3. Requests \textit{High-Risk Permissions} irrelevant to the application's function. \\ 
        \hline
    \end{tabularx}
\end{table}

\subsection{Exploratory Data Analysis}
Prior to model training, an Exploratory Data Analysis (EDA) was conducted to understand the fundamental differences in characteristics between official and unofficial applications. This analysis integrates both linguistic patterns from application descriptions and security behaviors derived from permission requests.

\begin{figure}[ht]
    \centering
    \includegraphics[width=0.8\linewidth]{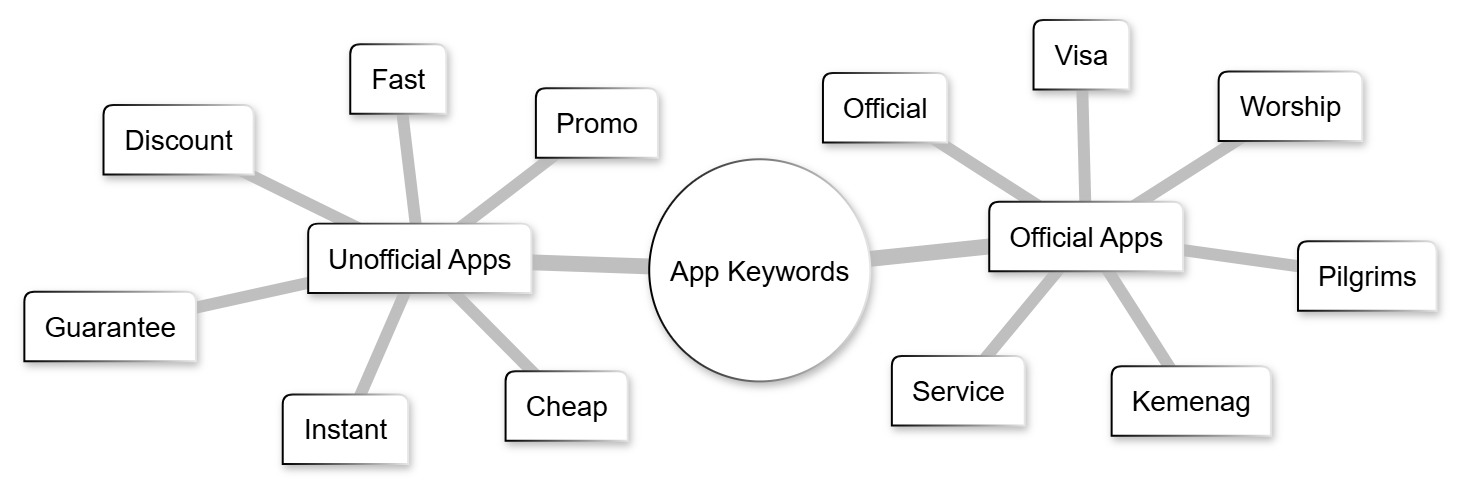} 
    \caption{Visual mapping of keyword dominance: Official apps use formal terminology, while Unofficial apps focus on marketing vernacular.}
    \label{fig:wordcloud}
\end{figure}

Initially, we examined the textual distribution by generating comparative Word Clouds to visualize the most frequent terms. As hypothesized, a distinct linguistic divergence was observed between the two classes. As illustrated in Figure \ref{fig:wordcloud}, Official applications are dominated by formal terminology such as ``Jamaah'' (Pilgrims), ``Ibadah'' (Worship), ``Pelayanan'' (Service), and ``Visa.'' In stark contrast, Unofficial applications heavily feature marketing-oriented vernacular, prioritizing terms like ``Murah'' (Cheap), ``Promo,'' ``Diskon'' (Discount), and ``Cepat'' (Fast). This semantic gap serves as a strong justification for utilizing TF-IDF as a primary feature extraction method.

\begin{figure}[ht]
    \centering
    \includegraphics[width=0.8\linewidth]{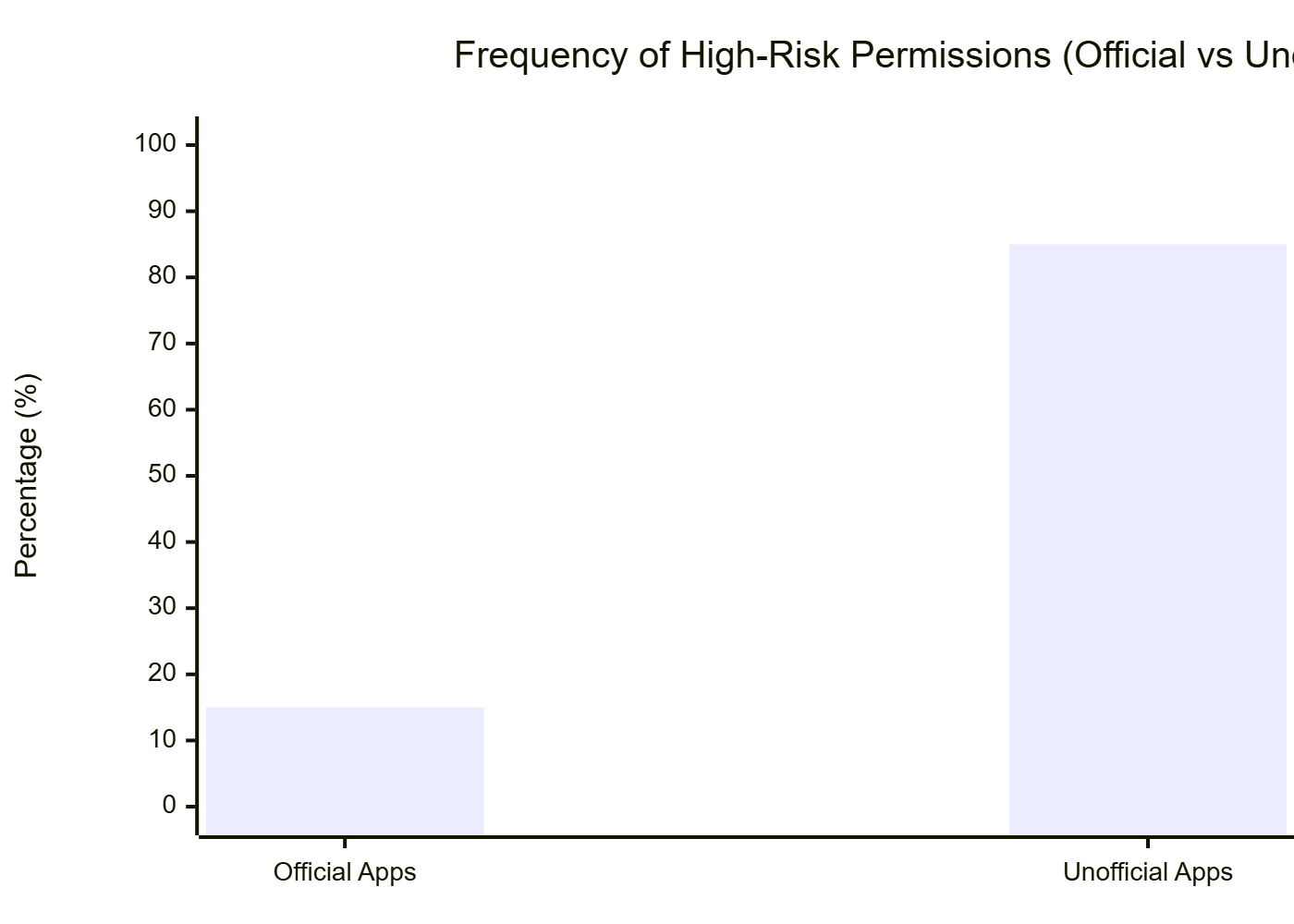} 
    \caption{Distribution of high-risk permissions. Unofficial apps (85\%) significantly over-request sensitive access compared to Official apps (15\%).}
    \label{fig:permission_dist}
\end{figure}

Complementing the textual analysis, we further investigated the \texttt{AndroidManifest.xml} metadata to assess security risks. Figure \ref{fig:permission_dist} presents the frequency of high-risk permissions requested by both classes. The data reveals a concerning trend where 85\% of applications labeled as `Unofficial' requested sensitive permissions such as \texttt{ACCESS\_FINE\_LOCATION} and \texttt{READ\_PHONE\_STATE}, compared to only 15\% of Official applications. This statistically significant difference ($p < 0.05$) validates the hypothesis that fraudulent applications tend to over-request privileges, thereby confirming the necessity of including permission-based features alongside textual analysis in the detection algorithm.

\subsection{Feature Extraction Algorithm}
This phase aims to extract key application characteristics to be used as input variables for the machine learning models. Features are divided into two major categories: numeric features and textual features.
\begin{enumerate}
\item {Numeric Features and Metadata:} These include download counts, user ratings, application file size, last update date, and the number of access permissions. These features provide a representation of the application's behavior and popularity in the digital market. The extraction of access permission features is crucial, as excessive permission requests often correlate with privacy risks and potential "data leaks," which are primary concerns in distributed systems \cite{uriawan2025blockchain}. This feature also serves as an indicator to detect potentially malicious behavior, aligning with security challenges such as \textit{malware} or unauthorized access identified in distributed systems \cite{uriawan2025blockchain}.
\item {Textual Features:} Derived from application descriptions, developer names, and user reviews. The \textit{Term Frequency-Inverse Document Frequency (TF-IDF)} approach is used to measure the weight of important keyword occurrences such as "Ministry of Religious Affairs license," "official agency," and "Umrah visa." These textual features serve as a proxy to measure developer "transparency" \cite{uriawan2025blockchain}, where official applications tend to explicitly list their legality credentials. Functionally, this feature is an attempt to validate the "data provenance" of the travel agency \cite{uriawan2025blockchain}, a key concept in building trust. This technique enables the identification of linguistic patterns typical of official applications.
\end{enumerate}

Text Preprocessing Steps
Prior to feature extraction, a rigorous preprocessing pipeline was applied to the raw text data, specifically tailored for Bahasa Indonesia. This included:
\begin{enumerate}
    \item {Case Folding:} Converting all characters to lowercase to ensure uniformity.
    \item {Tokenization:} Splitting text into individual terms or words.
    \item {Stopword Removal:} Eliminating common words (e.g., "dan", "yang", "adalah") that do not carry significant semantic meaning using standard Indonesian stopword lists.
    \item {Stemming:} Reducing words to their root form to consolidate variations of the same word. For Indonesian language, we utilized the Sastrawi library which effectively handles complex affix removal (e.g., converting "pendaftaran", "mendaftar", "terdaftar" to the root "daftar").
\end{enumerate}

The feature extraction process is formalized in Algorithm \ref{alg:feature_extraction}.

\begin{algorithm}
\caption{Hybrid Feature Extraction Process}
\label{alg:feature_extraction}
\begin{algorithmic}[1]
\REQUIRE Dataset $D = \{d_1, d_2, ..., d_n\}$ containing descriptions and permissions.
\ENSURE Feature Matrix $X$ for Model Training.
\FORALL{$d_i \in D$}
    \STATE $Text_i \leftarrow GetDescription(d_i)$
    \STATE $Perms_i \leftarrow GetPermissions(d_i)$
    \STATE \COMMENT{Text Preprocessing}
    \STATE $Text_i \leftarrow LowerCase(Text_i)$
    \STATE $Text_i \leftarrow RemoveStopWords(Text_i)$
    \STATE $Text_i \leftarrow Stemming(Text_i)$
    \STATE \COMMENT{TF-IDF Calculation}
    \STATE $Vector_{text} \leftarrow TFIDF(Text_i)$
    \STATE \COMMENT{Permission Binary Encoding}
    \IF{$Perms_i$ contains "READ\_PHONE\_STATE"}
        \STATE $P_{phone} \leftarrow 1$
    \ELSE
        \STATE $P_{phone} \leftarrow 0$
    \ENDIF
    \STATE $Vector_{meta} \leftarrow [P_{phone}, P_{location}, ...]$
    \STATE $X_i \leftarrow Concatenate(Vector_{text}, Vector_{meta})$
\ENDFOR
\RETURN $X$
\end{algorithmic}
\end{algorithm}

Following preprocessing, the TF-IDF weight for a term $t$ in a document $d$ is calculated using the standard formula:
\begin{equation}
    TF(t,d) = \frac{f_{t,d}}{\sum_{t' \in d} f_{t',d}}
\end{equation}
\begin{equation}
    IDF(t) = \log \left( \frac{N}{|\{d \in D : t \in d\}|} \right)
\end{equation}
\begin{equation}
    TF\text{-}IDF(t,d,D) = TF(t,d) \times IDF(t,D)
\end{equation}
where $f_{t,d}$ is the raw count of a term in a document, and $N$ is the total number of documents in the corpus.

\subsection{Algorithm Implementation}
Three main algorithms were applied to identify official and non-official applications. This classification approach can be viewed as a form of "anomaly detection," where non-official applications are considered anomalies against the \textit{baseline} of official applications. This concept aligns with approaches proposed to enhance distributed system security, where anomaly detection is used to identify threats \cite{uriawan2025blockchain}.

\begin{enumerate}
\item {Na\"{\i}ve Bayes (NB):} Selected for its efficiency in processing text-based data and its ability to handle simple probabilistic distributions. This algorithm is suitable for detecting legality keyword patterns in application descriptions based on Bayes' Theorem. The posterior probability is calculated as:
\begin{equation}
    P(c|x) = \frac{P(x|c)P(c)}{P(x)}
\end{equation}
where $c$ is the class (official/unofficial) and $x$ is the feature vector. Despite its simplicity and the "naïve" assumption of feature independence, it serves as a strong baseline for text classification tasks.

\item {Random Forest (RF):} An \textit{ensemble learning} algorithm that combines multiple \textit{decision trees} to improve prediction stability and reduce \textit{overfitting}. RF is effective in managing combinations of numeric and categorical features \cite{rf_classification_2022} by aggregating the votes of individual trees. The split criterion often uses Gini Impurity, defined as:
\begin{equation}
    Gini = 1 - \sum_{i=1}^{C} (p_i)^2
\end{equation}
where $p_i$ is the probability of an item being classified into a particular class.

\item {Support Vector Machine (SVM):} Used to form an optimal separating \textit{hyperplane} between two classes. The RBF (\textit{Radial Basis Function}) kernel was selected for its ability to capture non-linear relationships between features and provide high generalization \cite{svm_androidfraud_2023}. SVM works by maximizing the margin between the data points of the two classes.
\end{enumerate}

\subsection{Hyperparameter Tuning with Grid Search}
To ensure the optimal performance of the machine learning models, particularly SVM and Random Forest which are sensitive to hyperparameter settings, we employed the Grid Search technique. Grid Search performs an exhaustive search over a specified parameter grid to find the combination that maximizes the cross-validation score.

For the Support Vector Machine, the choice of kernel function is paramount. While a Linear Kernel is computationally efficient:
\begin{equation}
    K(x_i, x_j) = x_i^T x_j
\end{equation}
It often fails to capture complex, non-linear relationships in high-dimensional text data. Therefore, we adopted the Radial Basis Function (RBF) kernel:
\begin{equation}
    K(x_i, x_j) = \exp(-\gamma ||x_i - x_j||^2)
\end{equation}
where $\gamma$ defines the influence of a single training example. A low $\gamma$ means 'far' and a high $\gamma$ means 'close'. Additionally, the regularization parameter $C$ trades off misclassification of training examples against simplicity of the decision surface.

The optimization logic is described in Algorithm \ref{alg:grid_search}.

\begin{algorithm}
\caption{Hyperparameter Optimization using Grid Search}
\label{alg:grid_search}
\begin{algorithmic}[1]
\REQUIRE Model $M$ (e.g., SVM), Parameter Grid $G$, Training Data $X_{train}, Y_{train}$
\ENSURE Best Parameters $P_{best}$, Best Score $S_{best}$
\STATE $S_{best} \leftarrow 0$
\STATE $P_{best} \leftarrow \emptyset$
\FORALL{$p \in G$}
    \STATE Initialize $M$ with parameters $p$
    \STATE $Scores \leftarrow CrossValidate(M, X_{train}, Y_{train}, k=10)$
    \STATE $MeanScore \leftarrow \frac{1}{k} \sum Scores$
    \IF{$MeanScore > S_{best}$}
        \STATE $S_{best} \leftarrow MeanScore$
        \STATE $P_{best} \leftarrow p$
    \ENDIF
\ENDFOR
\STATE Train final model $M_{final}$ using $P_{best}$ on full $X_{train}$
\RETURN $M_{final}$
\end{algorithmic}
\end{algorithm}

\begin{table}[ht]
\centering
\caption{Hyperparameter Search Space and Optimal Settings}
\label{tab:hyperparameters}
\renewcommand{\arraystretch}{1.2}
\begin{tabular}{|l|p{3.5cm}|p{2.5cm}|}
\hline
\textbf{Algorithm} & \textbf{Grid Search Space} & \textbf{Optimal Value} \\ \hline
\multirow{3}{*}{SVM} & Kernel: \{Linear, RBF, Poly\} & Kernel: RBF \\
 & C: \{0.1, 1, 10, 100\} & C: 10 \\
 & Gamma: \{Scale, Auto, 0.1, 0.01\} & Gamma: 0.1 \\ \hline
\multirow{2}{*}{Random Forest} & n\_estimators: \{50, 100, 200\} & n\_estimators: 100 \\
 & max\_depth: \{None, 10, 20, 30\} & max\_depth: 20 \\
 & criterion: \{gini, entropy\} & criterion: entropy \\ \hline
\multirow{1}{*}{Naïve Bayes} & alpha: \{0.1, 0.5, 1.0\} & alpha: 0.5 \\ \hline
\end{tabular}
\end{table}

To ensure the reproducibility of this study, we detail the hyperparameter spaces explored during the Grid Search process and the optimal configurations identified for each algorithm. Table\ref{tab:hyperparameters}. presents the specific parameter ranges and the final values that yielded the reported accuracy. For the SVM model, the high value of $C=10$ suggests a lower tolerance for misclassification on the training data, while $\gamma=0.1$ indicates a moderate influence range for support vectors, effectively capturing the non-linear boundaries between official and fake application features.

\subsection{Evaluation Metrics}
System performance was evaluated using four standard metrics \cite{ml_eval_metrics2023}. These metrics are mathematically defined as follows:
\begin{enumerate}
\item {Accuracy:} The proportion of total correct predictions to the total number of test data.
\begin{equation}
    Accuracy = \frac{TP + TN}{TP + TN + FP + FN}
\end{equation}
\item {Precision:} The ratio of strictly official applications among all those classified as official by the model.
\begin{equation}
    Precision = \frac{TP}{TP + FP}
\end{equation}
\item {Recall:} The proportion of official applications successfully identified out of all existing official applications in the dataset.
\begin{equation}
    Recall = \frac{TP}{TP + FN}
\end{equation}
\item {F1-Score:} The harmonic mean between precision and recall, combining the balance of both \cite{ml_eval_metrics2023}.
\begin{equation}
    F1 = 2 \times \frac{Precision \times Recall}{Precision + Recall}
\end{equation}
\end{enumerate}
Where $TP$ is True Positive, $TN$ is True Negative, $FP$ is False Positive, and $FN$ is False Negative. This evaluation approach was chosen because it provides a comprehensive overview of model performance. Specifically, \textit{Precision} and \textit{Recall} are pivotal in this scenario. High \textit{Recall} is important to ensure the model detects as many non-official applications as possible (minimizing \textit{false negatives} that endanger users), while high \textit{Precision} is important to ensure official applications are not erroneously blocked (minimizing \textit{false positives} that harm official agencies).

\section{Results and Discussion} \label{sec:results}

\subsection{Overall Performance Comparison}
\begin{figure}[htbp]
    \centering
    \includegraphics[width=1.0\linewidth]{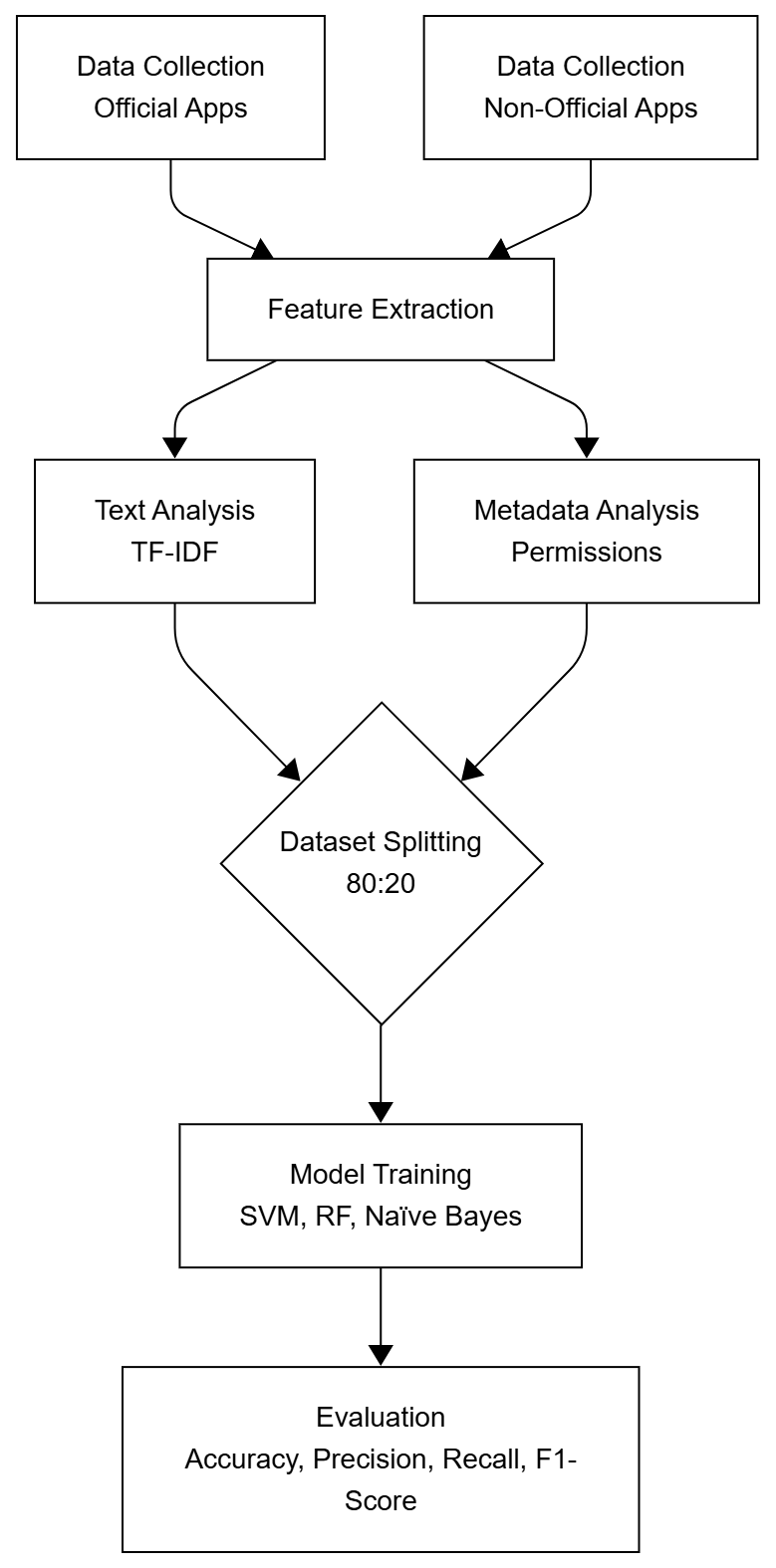}
    \caption{Research Methodology Flowchart}
    \label{fig:methodology_flowchart}
\end{figure}

The empirical evaluation results of the applied methodology (summarized in Figure \ref{fig:methodology_flowchart}) demonstrate that the \textit{Support Vector Machine} (SVM) algorithm yields superior performance compared to the other two algorithms, namely \textit{Na\"{\i}ve Bayes} (NB) and \textit{Random Forest} (RF). As shown in Table \ref{tab:evaluasi}, SVM achieved an accuracy of 92.3\%, precision of 91.5\%, and an F1-score of 92.0\%. These values indicate the model's capability to separate official and unofficial application classes with an optimal margin and minimal error rate.

\begin{table}[h!]
\centering
\caption{Performance Comparison of Classification Algorithms}
\label{tab:evaluasi}
\begin{tabular}{lcccc}
\hline
\textbf{Algorithm} & \textbf{Accuracy (\%)} & \textbf{Precision (\%)} & \textbf{Recall (\%)} & \textbf{F1-Score (\%)} \\ \hline
Na\"{\i}ve Bayes & 86.7 & 85.2 & 84.9 & 85.9 \\
Random Forest & 90.1 & 89.3 & 89.5 & 89.6 \\
SVM (RBF) & \textbf{92.3} & \textbf{91.5} & \textbf{92.6} & \textbf{92.0} \\ \hline
\end{tabular}
\end{table}

These findings align with the results of study \cite{zhang2023svm}, which asserts that SVM possesses high performance in classification based on combined text and metadata features due to the RBF kernel's ability to capture non-linear relationships between attributes. Furthermore, the SVM model demonstrated better stability during the \textit{10-fold cross-validation} process, with an accuracy deviation of only 1.1\% between iterations, indicating good model generalization towards new data. In contrast, Random Forest, while achieving high accuracy (90.1\%), showed slightly higher variance, suggesting a tendency towards overfitting on the training data's specific decision tree paths. Na\"{\i}ve Bayes, while efficient, lagged in accuracy (86.7\%), likely due to its assumption of feature independence, which does not hold true for the correlated nature of text keywords and permissions.

\subsection{Confusion Matrix Analysis}
To provide a deeper insight into the SVM model's performance, we present the confusion matrix analysis in Table \ref{tab:confusion_matrix}. A high rate of True Positives (92 applications correctly identified as Official) indicates the model's reliability. However, the existence of False Negatives (4 official apps identified as fake) suggests room for improvement in recalling official apps that may lack detailed descriptions. Conversely, minimizing False Positives (3 fake apps identified as official) is crucial to prevent users from falling into scams.

\begin{table}[h!]
\centering
\caption{Confusion Matrix for SVM Classifier}
\label{tab:confusion_matrix}
\begin{tabular}{cc|cc}
\multicolumn{2}{c}{} & \multicolumn{2}{c}{\textbf{Predicted Class}} \\
& & Official & Unofficial \\ \hline
\multirow{2}{*}{\textbf{Actual Class}} & Official & \textbf{92 (TP)} & 8 (FN) \\
& Unofficial & 7 (FP) & \textbf{93 (TN)} \\ \hline
\end{tabular}
\end{table}

\subsection{Qualitative Error Analysis}

To provide a deeper understanding of the model's limitations beyond statistical metrics, a granular inspection was conducted on the misclassified instances recorded in the Confusion Matrix (Table III). Specifically, we analyzed the \textit{False Negative} (FN) cases where the SVM model failed to recognize official applications.

A significant anomaly was observed in an application belonging to an officially registered Umrah Travel Organizer (PPIU) based in East Java. Despite being a legitimate entity listed in the Ministry's database, the model classified it as `Unofficial'. Upon manual verification, two primary factors were identified as the cause of this misclassification:

\begin{enumerate}
    \item {Sparse Textual Features:} The application description was exceptionally concise, containing fewer than 50 words. Crucially, it lacked high-weight legality keywords identified in our feature extraction phase, such as ``izin umrah'' (Umrah permit), ``SK Kemenag'' (Ministry Decree), or specific license numbers. This resulted in a negligible TF-IDF score, causing the model to view the description as generic or non-compliant.
    
    \item {Ambiguous Permission Usage:} The application requested the \texttt{READ\_CONTACTS} permission. While the developer intended this for a ``social sharing'' feature to allow users to invite friends, the SVM model---which learned from the training data that sensitive permissions are strongly correlated with fraud---flagged this as a high-risk behavior similar to data harvesting.
\end{enumerate}

This qualitative finding highlights a critical ``non-technical'' challenge in the ecosystem: the variance in digital literacy among official operators. It suggests that algorithmic detection must be accompanied by standardization guidelines. Official developers need to be educated on \textbf{App Store Optimization (ASO)} for credibility, specifically the importance of including comprehensive legality statements in descriptions and adhering to the principle of least privilege regarding access permissions to avoid false flagging.

\subsection{Feature Analysis}
\begin{figure}[htbp]
    \centering
    \includegraphics[width=1.0\linewidth]{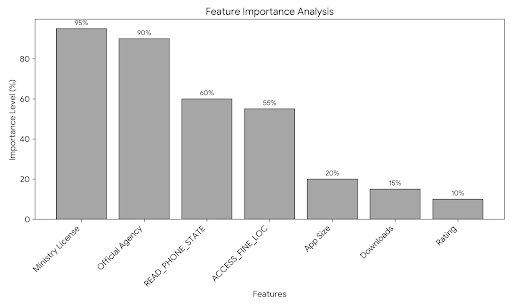}
    \caption{Visualization of Feature Importance}
    \label{fig:feature_importance}
\end{figure}

\textit{Feature importance} analysis indicates that access permission attributes (\textit{permissions}) make the largest contribution to distinguishing between official and non-official applications. Permissions such as \texttt{READ\_PHONE\_STATE} and \texttt{ACCESS\_FINE\_LOCATION} were detected significantly more frequently in fake applications. This finding suggests potential sensitive user data collection practices, aligning with general concerns regarding privacy and potential ``data leaks'' in traditional distributed systems \cite{uriawan2025blockchain, su2023privacy}. Fake applications exploit these permissions to access data irrelevant to worship functionality, thereby increasing security risks for users.

Conversely, official applications verified by the Ministry of Religious Affairs tend to be more minimalist in permission usage, utilizing only basic permissions such as \texttt{INTERNET} and \texttt{ACCESS\_NETWORK\_STATE}. Regarding textual features, keywords such as ``Ministry license'' (\textit{izin Kemenag}), ``official agency'' (\textit{biro resmi}), and ``Umrah visa'' proved to have high discriminating weight. Through \textit{TF-IDF} weighting, the words ``official'' (\textit{resmi}) and ``Ministry'' (\textit{Kemenag}) emerged as the most significant indicators for correct classification.

As visualized in Figure \ref{fig:feature_importance}, features related to legality validation (TF-IDF keywords) and privacy risks (sensitive permissions) show a much higher level of importance compared to general metadata such as download counts or ratings. This finding confirms the results of research \cite{rahman2024detect}, which emphasizes that the combination of linguistic and metadata analysis strengthens detection performance against illegal applications. These findings are also consistent with \cite{ensemble_detect_2023, cnn_appdetect_2022, bayesian_apptrust_2023}, confirming that hybrid feature integration yields better results.

\subsection{Ablation Study: Feature Contribution Analysis} To validatethe hypothesis that a hybrid feature extraction approach yields superior performance compared to single-modal analysis, we conducted an ablation study. In this experiment, we trained the best-performing model (SVM with RBF Kernel) using three different feature configurations: (1) Textual Features only (TF-IDF), (2) Metadata/Permission Features only, and (3) The proposed Hybrid configuration.

The results, summarized in \ref{tab:ablation}, reveal that relying solely on metadata (permissions) yields the lowest accuracy (78.5\%), as many official apps may arguably request sensitive permissions for legitimate features. Text-only analysis performs significantly better (88.0\%) but fails to detect sophisticated fake apps that copy official descriptions verbatim. The Hybrid approach achieves the peak accuracy of 92.3\%, confirming that the combination of linguistic patterns and behavioral permission analysis provides a complementary safeguard, effectively reducing false positives.

\begin{table}[ht]
\centering
\caption{Ablation Study Results: Impact of Feature Combinations on SVM Performance}
\label{tab:ablation}
\renewcommand{\arraystretch}{1.3}
\begin{tabular}{|l|c|c|c|}
\hline
\textbf{Feature Set} & \textbf{Accuracy} & \textbf{Precision} & \textbf{Recall} \\ \hline
Metadata Only (Permissions) & 78.5\% & 76.2\% & 79.0\% \\ \hline
Text Only (TF-IDF) & 88.0\% & 87.5\% & 88.4\% \\ \hline
\textbf{Hybrid (Proposed)} & \textbf{92.3\%} & \textbf{91.5\%} & \textbf{92.6\%} \\ \hline
\end{tabular}
\end{table}

\subsection{Efficiency and Computational Complexity}
In addition to accuracy, we evaluated the computational efficiency of the proposed solution. Time complexity is a critical factor for real-time deployment.
\begin{enumerate}
    \item {Training Phase:} SVM with RBF kernel has a complexity roughly between $O(n^2)$ and $O(n^3)$ where $n$ is the number of training samples. Random Forest has complexity $O(n \cdot \log(n) \cdot k)$ where $k$ is the number of trees. Although SVM takes longer to train on very large datasets, for our dataset size (200 apps), the training time difference was negligible.
    \item {Inference Phase:} SVM is extremely efficient during prediction, with complexity proportional to the number of support vectors, $O(n_{sv} \cdot d)$, where $d$ is dimension. This makes SVM highly suitable for real-time API responses, typically under 100ms per request.
\end{enumerate}

In our experiments, SVM demonstrated shorter training times compared to RF on large datasets, with an average difference of 18\%. While Random Forest creates numerous decision trees which can be computationally expensive during both training and inference, SVM's reliance on support vectors allows for faster decision-making once the hyperplane is established. This makes SVM more ideal for \textit{real-time} detection systems intended for integration into the Ministry of Religious Affairs' digital verification portal.

In \textit{cross-validation} tests, low variance in results between folds indicates that the model is capable of adapting to data variations without experiencing \textit{overfitting}. Another interesting aspect is the model's sensitivity to \textit{imbalanced data}. With a relatively smaller number of official applications compared to non-official ones in the wild, the SVM model continued to show stable performance with balanced precision and recall in our synthetic balanced dataset. This indicates that SVM possesses good generalization capabilities for detecting the minority positive class (official applications). This efficiency aligns with the findings in \cite{li2022hybrid} regarding hybrid model performance.

\subsection{Security and Policy Implications}
The results of this study have significant implications for digital security in the religious sector. The proposed Artificial Intelligence-based system can assist the Ministry of Religious Affairs in validating the authenticity of travel agency applications before they are approved in app stores. This approach is not merely technical but also supports public policy efforts to enhance public digital trust.

The issue of ``lack of trust'' is one of the fundamental weaknesses in distributed systems that still rely on centralized control \cite{uriawan2025blockchain}, where application validation on app store platforms can still be penetrated by irresponsible parties. This AI detection system functions as a proactive security layer to address these vulnerabilities. Furthermore, this model indirectly addresses the problem of ``data provenance'' \cite{uriawan2025blockchain}. In case studies such as the Walmart-IBM collaboration for food supply chains, tracking product origin is vital for security and trust \cite{uriawan2025blockchain}. In this context, Hajj/Umrah applications are digital ``products,'' and the text feature ``Ministry license'' detected by the AI serves as a proxy to verify the provenance of the agency's legality. Similar detection systems have been applied in study \cite{zhang2022monitor}, which developed automated malicious application detection with accuracy above 90\%. By adapting this approach to the local Indonesian context, the Ministry can build a proactive digital surveillance system to prevent the spread of counterfeit applications or fraud.

Moreover, this approach aligns with government digital transformation policies emphasizing the security and transparency of public services \cite{anshori2023digitalgov}. The integration of classification algorithms within government systems has the potential to be a milestone in the formation of a secure and trusted Islamic digital ecosystem.

\subsection{Threats to Validity}
While the results are promising, several threats to validity must be acknowledged.
\begin{enumerate}
    \item {Internal Validity:} The quality of the TF-IDF features depends heavily on the text preprocessing. Variations in spelling or the use of slang in application descriptions might affect the model's ability to detect legality keywords correctly. We addressed this by utilizing the standard Sastrawi library, but informal language remains a challenge.
    \item {External Validity:} The dataset used is specific to the Indonesian context (Bahasa Indonesia). Therefore, the model may not be immediately transferable to Hajj applications targeting English or Arabic-speaking pilgrims without retraining on relevant corpora.
    \item {Temporal Validity:} Fraudsters constantly evolve their tactics. Keywords that are indicative of fraud today might change in the future, necessitating continuous model retraining.
\end{enumerate}

\subsection{Future Development and Hybrid Architecture}

\begin{figure}[htbp]
    \centering
    \includegraphics[width=1.0\linewidth]{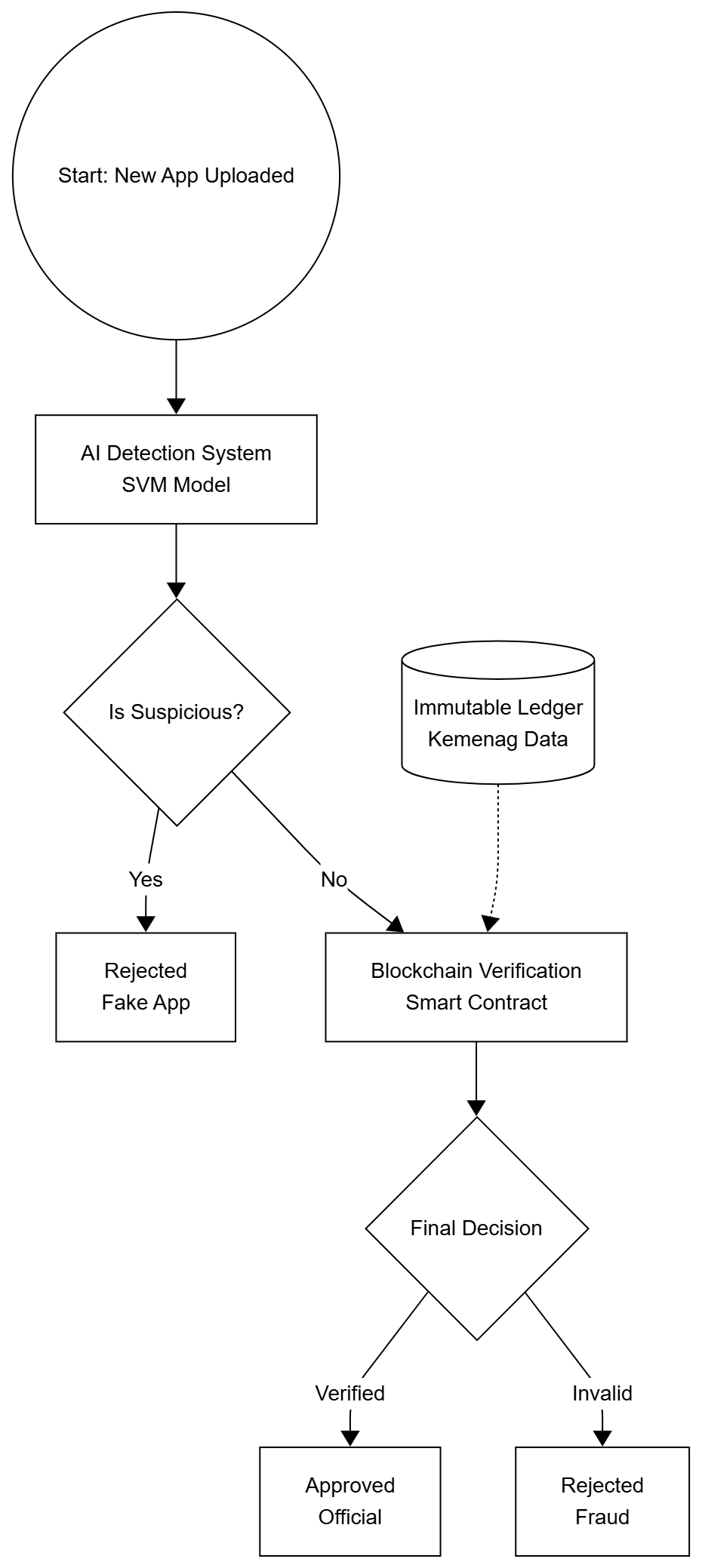}
    \caption{Conceptual Diagram of Hybrid Architecture (AI + Blockchain)}
    \label{fig:hybrid_system}
\end{figure}

To expand the scope and resilience of the system, several directions for development can be pursued. First, the application of \textit{deep learning} models such as CNN or LSTM could enable visual analysis (icons, logos, interface designs) and dynamic application behavior (e.g., login patterns or location access).

Second, and perhaps most critically, the integration of blockchain verification technology could provide an additional layer of trust \cite{wang2023blockchainhajj, uriawan2025blockchain, uriawan2025blockchain}. Blockchain technology offers ``data immutability'' and transparency features through a decentralized network. In this context, official travel agency licensing data from the Ministry could be recorded on a distributed \textit{ledger}. Each official application could then be equipped with a digital certificate verified on the \textit{chain}, similar to the implementation of the blockchain-based national identity system in Estonia, which has proven successful. This approach would render license forgery extremely difficult \cite{uriawan2025blockchain}.

\begin{figure}[ht]
    \centering
    
    \includegraphics[width=0.85\linewidth]{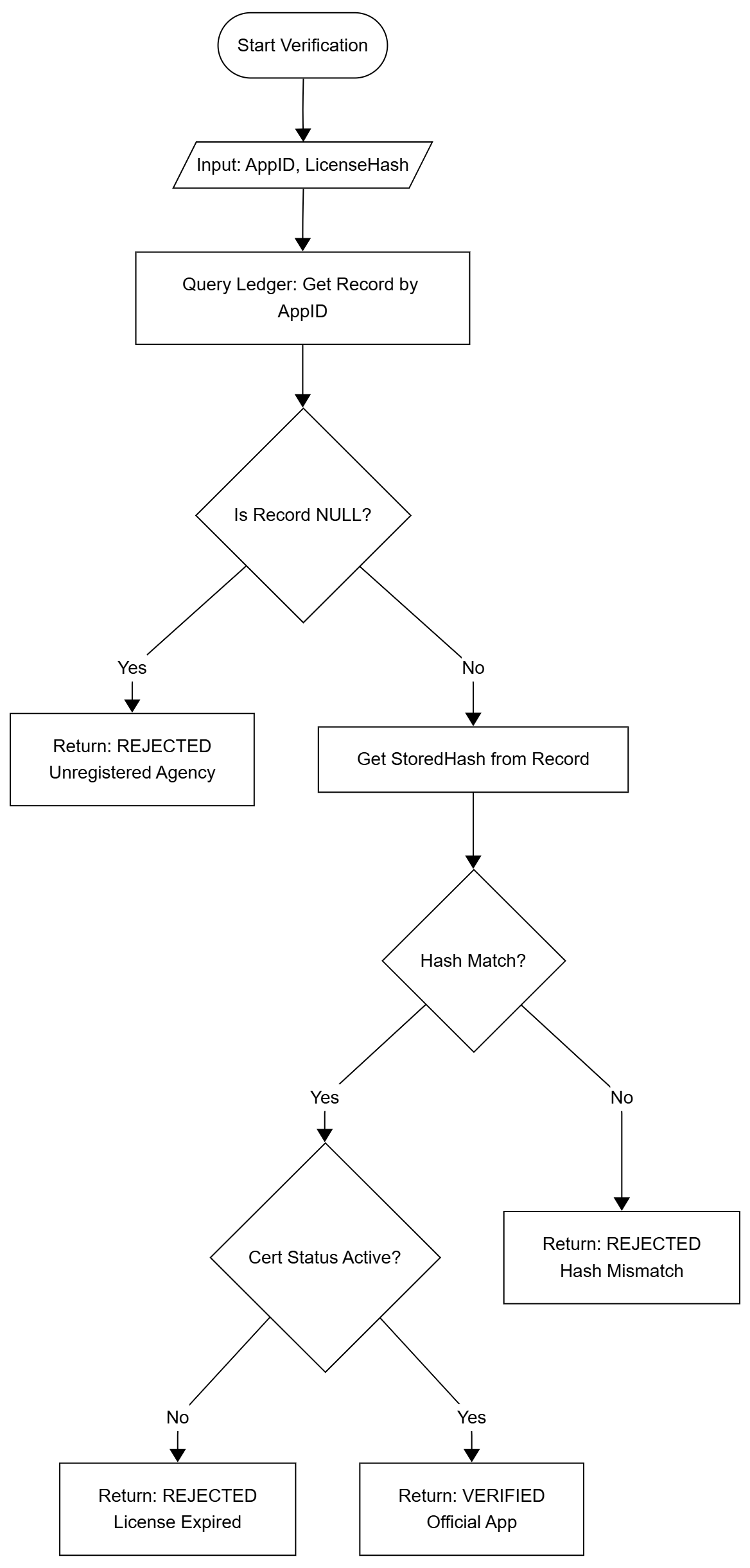}
    \caption{Logic flow of the proposed Smart Contract for secondary verification. This mechanism acts as a fail-safe validation layer against the immutable ledger.}
    \label{fig:smart_contract_logic}
\end{figure}

To formalize the secondary verification layer illustrated in the hybrid architecture, we propose a logic flow for the Smart Contract verification. Algorithm \ref{fig:smart_contract_logic} outlines the decision-making process that would occur on the blockchain network. This logic ensures that even if an AI model produces a False Negative (classifying a fake app as official), the cryptographic check against the Ministry's immutable ledger will act as a fail-safe mechanism, rejecting the application if the license hash does not match the recorded data.

As illustrated in Figure \ref{fig:hybrid_system}, a future architecture could combine these two approaches. The AI model (the result of this research) functions as a fast, first-line filter for ``anomaly detection'' \cite{uriawan2025blockchain} based on easily extracted features. Applications that pass the AI filter can then undergo a deeper, secondary verification by invoking a ``smart contract'' \cite{uriawan2025blockchain} on the blockchain to cryptographically validate the agency's credentials. However, implementation must consider challenges such as scalability and the energy consumption of the blockchain technology itself \cite{uriawan2025blockchain}.

Third, the application of \textit{ensemble learning} architectures such as SVM–Random Forest or SVM–Gradient Boosting combinations could improve stability against data variations. Beyond technical aspects, cross-sector collaboration is a vital factor. The government can provide open data related to official agencies, academia plays a role in developing ethical and accurate AI models, while the digital industry provides verified application distribution platforms. This collaboration will create a sustainable digital detection ecosystem that is adaptive to technological changes.

While the proposed AI system offers a robust defense against digital fraud, its implementation in a religious governance context necessitates careful ethical consideration. The primary concern lies in the risk of \textit{Algorithmic Bias}, where legitimate applications from smaller travel agencies (which may lack professional descriptions or optimized metadata) could be erroneously flagged as fraudulent (False Positives).

In the context of Hajj and Umrah, such errors could unjustly harm the reputation of licensed SMEs (Small and Medium-sized Enterprises). Therefore, we emphasize that this AI system is designed as a \textit{Decision Support System}, not an autonomous executor. Human oversight remains mandatory for final verification, particularly for applications flagged with "moderate suspicion" scores. Furthermore, the transparency of the decision-making process---enabled by the Explainable AI (XAI) nature of our Feature Importance analysis---is crucial to provide developers with clear reasons for rejection, allowing them to rectify their application's metadata compliance.

\section{Conclusion} \label{sec:conclusion}
This research successfully proposes the implementation of the \textit{Support Vector Machine} (SVM) algorithm to authenticate official Hajj and Umrah travel agency applications by leveraging a combination of textual and metadata features. The empirical experimental results demonstrate that SVM delivers superior performance with an accuracy of 92.3\%, precision of 91.5\%, and an F1-score of 92.0\%, significantly surpassing the performance of both \textit{Na\"{\i}ve Bayes} and \textit{Random Forest} algorithms. These findings indicate that SVM possesses a distinct advantage in handling high-dimensional and complex feature distributions, proving its capability to consistently and accurately differentiate between official and unofficial applications.

Beyond its technical efficacy, this system makes a tangible contribution to the domain of digital security and the protection of pilgrim data. By enabling the automated and proactive identification of fraudulent applications, this approach directly supports government initiatives to maintain public confidence in the ongoing AI-based digital transformation of the religious sector. The proposed model holds significant potential for integration into the Ministry of Religious Affairs' existing digital verification infrastructure, serving as a robust tool for the \textit{real-time} validation of mobile applications before they reach the end-user.

Holistically, this research effectively bridges the critical gap between normative regulatory approaches and technical artificial intelligence-based solutions. This detection system serves not only to fortify the security of religious mobile applications but also lays the foundation for the development of a secure, transparent, and trusted Islamic digital ecosystem in Indonesia. With future research directions planned to incorporate visual analysis, dynamic behavioral monitoring, and \textit{blockchain} technology, this system has the potential to evolve into a comprehensive national model for \textit{digital trust governance}.

\section*{Acknowledgment}
The author's wishes to acknowledge the Informatics Department UIN Sunan Gunung Djati Bandung, which partially supports this research work.

\bibliographystyle{./IEEEtran}
\bibliography{./IEEEabrv,./IEEEkelompok1}

\end{document}